%% file: egpaper.tex
\documentclass[10pt,twocolumn,letterpaper]{article}
\usepackage{wacv}
\usepackage{times}
\usepackage{epsfig}
\usepackage{graphicx}
\usepackage{amsmath}
\usepackage{amssymb}
\usepackage{booktabs}
\usepackage[accsupp]{axessibility}

\usepackage{enumitem}
\usepackage[numbers,sort,compress]{natbib}
\usepackage{siunitx}
\usepackage{tabularx}
\usepackage{multirow}
\usepackage{caption} 

\usepackage[dvipsnames, table]{xcolor} 
\usepackage{glossaries}
\usepackage{pifont}
\newcommand{\xmark}{\ding{55}}%
\usepackage[symbol]{footmisc}

\include{math_includes}

\def\wacvPaperID{****} 
\wacvalgorithmstrack   
\wacvfinalcopy 

\ifwacvfinal
\usepackage[pagebackref=true,breaklinks=true,colorlinks,bookmarks=false]{hyperref} 
\else
\usepackage[pagebackref=true,breaklinks=true,colorlinks,bookmarks=false]{hyperref}
\fi

\newcommand{\projecturl}{\url{https://mm4spa.github.io/tvcalib}}
\usepackage[capitalise]{cleveref} 
\Crefname{equation}{Eq.}{Eqs.}
\Crefname{section}{Sec.}{Sec.}
\Crefname{appendix}{Appx.}{Appx.}
\newacronym{method}{\emph{TVCalib}}{\emph{TVCalib}}
\newacronym{pnp}{P\emph{n}P}{Perspective-$n$-Point}
\newacronym{pnpf}{P\emph{n}P(f)}{perspective-n-point}
\newacronym{dlt}{DLT}{Direct Linear Transform}
\newacronym{ndc}{\emph{NDC}}{Normalized Device Coordinates}
\newacronym{fov}{\emph{FoV}}{Field of View}
\newacronym{stn}{STN}{Spatial Transformer Network}
\newacronym{dof}{DoF}{Degree of Freedom}
\newacronym{iou}{$IoU$}{Intersection over Union}
\newacronym{sgd}{SGD}{Stochastic Gradient Descent}
\newacronym{ransac}{RANSAC}{Random Sample Consensus}
\newacronym{cnn}{CNN}{Convolutional Neural Network}

\begin{document}

\title{TVCalib: Camera Calibration for Sports Field Registration in Soccer\\{\normalsize\projecturl}}
\author{
Jonas Theiner$^1$
\and
Ralph Ewerth$^{1,2}$
\and
$^1$~\normalsize L3S Research Center, Leibniz University Hannover, Hannover, Germany\\
$^2$~\normalsize TIB – Leibniz Information Centre for Science and Technology, Hannover, Germany\\
{\tt\small theiner@l3s.de \quad ralph.ewerth@tib.eu}
}

\thispagestyle{empty}
\twocolumn[{%
\renewcommand\twocolumn[1][]{#1}%
\maketitle

\begin{center}
    \centering
    \captionsetup{type=figure}
    \includegraphics[width=\textwidth]{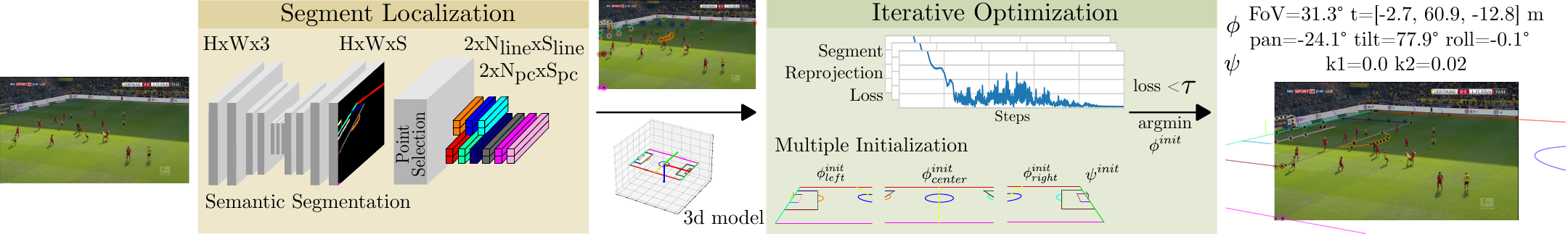}
    \captionof{figure}{Our proposed framework for 3D sports field registration:~(1)~\emph{segment localization} performs instance segmentation and selects appropriate points with respective label from a known calibration object~(3D model), and~(2)~our main contribution, the calibration module, which predicts camera parameters~$\phi$ by iteratively minimizing the \emph{segment reprojection loss}.}
    \label{fig:workflow}
\end{center}%
}]

\begin{abstract}
Sports field registration in broadcast videos is typically interpreted as the task of homography estimation, which provides a mapping between a planar field and the corresponding visible area of the image. In contrast to previous approaches, we consider the task as a camera calibration problem. First, we introduce a differentiable objective function that is able to learn the camera pose and focal length from segment correspondences~(e.g.,~lines, point clouds), based on pixel-level annotations for segments of a known calibration object. The calibration module iteratively minimizes the segment reprojection error induced by the estimated camera parameters. Second, we propose a novel approach for 3D~sports field registration from broadcast soccer images. Compared to the typical solution, which subsequently refines an initial estimation, our solution does it in one step. The proposed method is evaluated for sports field registration on two datasets and achieves superior results compared to two state-of-the-art approaches.
\end{abstract}

\section{Introduction}
%
%
Camera calibration is fundamental for numerous computer vision applications such as tracking, autonomous driving, robotics, augmented reality, etc.
Existing literature has extensively studied this problem for fully calibrated, partially calibrated, and uncalibrated cameras in various settings~\cite{jeong2021self}, for different types of data~(e.g. monocular images, image sequences, RGB-D images, etc.), and related tasks like 3D~reconstruction. 
%
%
Broadcast videos of sports events are a widely available data source. 
The ability to calibrate from a single, moving camera with unknown and changing camera parameters enables various augmented reality~\cite{d2010review} and sports analytics applications~\cite{cuevas2020techniques, Sha2018InteractiveSportsAnalytics}.

%
%
%
The sports field serves as a calibration object~(known dimensions according to the game rules). However, the non-visibility of appropriate keypoints in broadcast soccer videos~\cite{Citraro2020Realtimecamera}
and the unknown focal length prevent a sufficiently accurate direct computation of a homography or intrinsics and extrinsics from 2D-3D~(keypoint)~correspondences~\cite{Zhang2000flexiblenewtechnique, Harltey2006Multipleviewgeometry, Alem1978BIOMECHANICSAPPLICATIONSDIRECT, Zheng2014GeneralSimpleMethod}.
%
%
It has been shown that line~\cite{Homayounfar2017SportsFieldLocalization, Nie2021RobustEfficientFramework}, area~\cite{Sha2020EndEndCamera, Nie2021RobustEfficientFramework, chen2019sports}, point features with additional information~\cite{Nie2021RobustEfficientFramework, chu2022sports, Citraro2020Realtimecamera} are more suitable for accurate sports field registration. 
Previous approaches~\cite{Shi2022SelfSupervisedShape, chu2022sports, Nie2021RobustEfficientFramework, Sha2020EndEndCamera, Jiang2020OptimizingLearnedErrors, chen2019sports} treat the task as homography estimation instead of calibration despite the estimation of camera parameters enables further applications~(e.g., virtual stadiums, automatic camera control, or offside detection). 
To date, homography-based approaches may provide camera parameters for a first coarse initial estimation, but the more accurate results are usually based on homography refinements.

%
%
In this paper, we suggest to consider sports field registration as a calibration task
and estimate individual camera parameters~(position, rotation, and focal length) of the standard pinhole camera model~(and potential radial lens distortion coefficients) from an image without relying on keypoint correspondences between the image and 3D~scene. 
%
%
Contrary to the dominant direction of first estimating an initial result and then refining it, our method does both in one step without relying on training data for the calibration part.
Further, we use a dense representation of the visible field, i.e., directly leverage a small fraction of labeled pixel representing field segments instead of a~(deep) image representation for both initial estimation~\cite{chen2019sports, Sha2020EndEndCamera} or refinement~\cite{Sha2020EndEndCamera, Nie2021RobustEfficientFramework, chen2019sports,  Jiang2020OptimizingLearnedErrors, Shi2022SelfSupervisedShape, chu2022sports, Citraro2020Realtimecamera}.

%
%
We propose~(1)~a generic differentiable objective function that exploits the underlying primitives of a 3D object and measures its reprojection error.
We additionally suggest~(2)~a novel framework for 3D sports field registration~(\acrshort{method}) from TV~broadcast frames~(Fig.~\ref{fig:workflow}), 
including semantic segmentation, point selection, the calibration module, and result verification, where the calibration module iteratively minimizes the \emph{segment reprojection loss}.
%
%
The effectiveness of our method is evaluated on two real-world soccer broadcast datasets~(\emph{SoccerNet-Calibration}~\cite{cioppa2022soccernet} and \emph{World Cup 2014}~(WC14)~\cite{Homayounfar2017SportsFieldLocalization}), and we compare to state of the art in~2D~sports field registration.

The rest of the paper is organized as follows. 
\cref{sec:rw} provides an overview on~2D~sports field registration and the related calibration task.
In \cref{sec:method}, we describe the proposed \acrshort{method} in detail. 
Experimental results and a comparison with the state of the art are reported in \cref{sec:exp}, while \cref{sec:conclusions} concludes the paper and outlines areas of future work.

\section{Related Work on Sports Field Registration}\label{sec:rw}

Common to most approaches for sports field registration is that they predict homography matrices from main broadcast videos in team sports while the focus is on soccer.
Early approaches rely on local feature matching in combination with \acrfull{dlt} for homography estimation~\cite{Carr2012Pointlesscalibration, Gupta2011UsingLineEllipse, Ghanem2012Robustvideoregistration, puwein2011robust}, and both line and ellipse features are already used~(e.g., \cite{Gupta2011UsingLineEllipse, Nie2021RobustEfficientFramework, Homayounfar2017SportsFieldLocalization, Sha2020EndEndCamera}).
More recent approaches rely on learning a representation of the visible sports field by performing different variants of semantic segmentation. 
Approaches directly predict or regress an initial homography matrix~\cite{Nie2021RobustEfficientFramework, Jiang2020OptimizingLearnedErrors, Shi2022SelfSupervisedShape, chu2022sports} or search for the best matching homography in a reference database~\cite{Sha2020EndEndCamera, chen2019sports, Sharma2018AutomatedTopView, zhang2021high} containing synthetic images with known homography matrices or camera parameters.
This estimation is called initial estimation~$\hat\mH_{\mathit{init}}$ which is subsequently refined by the majority of approaches and considered as the relative \mbox{(non-)affine} image transformation~$\hat\mH_{\mathit{rel}}$ between the segmented input image and the predicted or retrieved image, finally resulting in $\hat\mH = \hat\mH_{\mathit{init}}\hat\mH_{\mathit{rel}}\in\sR^{3\times3}$.

Next, we review existing approaches regarding segmentation, initial estimation, refinement, and finally discuss how to access camera parameters.

\paragraph{Semantic Segmentation:}\label{rw:segmentation}
Some approaches use hand-crafted methods to detect lines, edges, ellipses, vanishing points~(lines)~or to perform area segmentation~(see~\cite{hayet2005fast, Cuevas2020Automaticsoccerfield} for an overview). 
Convolutional Neural Networks with increased receptive field~(e.g., via dilated convolutions~\cite{chen2017rethinking} or non-local blocks~\cite{wang2018non}) are used perform various types of image segmentation tasks, e.g., keypoint prediction, line segmentation, or area masking.
\citet{chen2019sports} first remove the background and then predict a binary mask representing all field markings.
\citet{Homayounfar2017SportsFieldLocalization} predict points from specific line and circle segments. 
Other approaches segment the sports field into four different areas~\cite{Sha2020EndEndCamera}, or detect appropriate field keypoints and player positions~\cite{Citraro2020Realtimecamera}.
\citet{Nie2021RobustEfficientFramework} aim to learn a strong field representation by jointly predicting uniformly sampled grid points, line features, and area features.
Inspired by predicting a dense grid of points~\cite{Nie2021RobustEfficientFramework}, \citet{chu2022sports} formulate the task as an instance segmentation problem. 
We also apply instance segmentation~\cite{chu2022sports} but on all individual field segments.

\paragraph{Initial Estimation:}\label{rw:initial_estimation}
A grid of uniformly sampled and predicted points~\cite{Nie2021RobustEfficientFramework,chu2022sports} or predicted keypoints~\cite{Citraro2020Realtimecamera, Cuevas2020Automaticsoccerfield} is the input for \acrshort{dlt}~(and variants)~\cite{Harltey2006Multipleviewgeometry} to get usually a rough initial homography estimation.
Segmented~\cite{Jiang2020OptimizingLearnedErrors} or raw~\cite{Shi2022SelfSupervisedShape} images are used to directly predict the homography or to regress four points. Still, such approaches require annotated homography matrices for training~\cite{Nie2021RobustEfficientFramework}.
\citet{Sharma2018AutomatedTopView} develop a large synthetic dataset of camera poses, whereby~\citet{chen2019sports} train a Siamese network to learn a representation of the respective segmentation mask and retrieve the nearest neighbor given an input mask.
\citet{Sha2020EndEndCamera} use a much smaller database and consequently leave the refinement module to perform large non-affine transformations to the semantic input image.

\paragraph{Homography Refinement:}\label{rw:refinement}
Homography refinement is a crucial step in order to obtain a more accurate estimate, if necessary~\cite{chu2022sports}. 
Previous approaches~\cite{chen2019sports, zhang2021high} use the Lucas-Kanade algorithm~\cite{baker2004lucas}, also in combination with spatial pyramids~\cite{Ghanem2012Robustvideoregistration} with the assumption that the image transformation is small.
To handle large non-affine transformations, the \acrfull{stn}~\cite{jaderberg2015spatial} was introduced in sports field registration.
Refinement is performed during one feed-forward step~\cite{Sha2020EndEndCamera} or by iteratively minimizing the difference between the input image and the initial estimation~\cite{Nie2021RobustEfficientFramework, Jiang2020OptimizingLearnedErrors}.

\paragraph{Accessing Individual Camera Parameters:}\label{rw:accessing_from_homograpgy}
\citet{Carr2012Pointlesscalibration} leverage a gradient-based image alignment algorithm to estimate camera and lens distortion parameters, but the refinement is performed on the homography.
A database of synthetic templates~\cite{chen2019sports, Sha2020EndEndCamera} allows for direct access to the camera pose as projective geometry is used to create template images. However, the smaller the database, the larger the reprojection error is without a refinement step.
Despite the focus on homographies, it allows us to access individual camera parameters, at least with homography decomposition~\cite{Harltey2006Multipleviewgeometry, Citraro2020Realtimecamera}. 
\citet{Citraro2020Realtimecamera} decompose the initial estimated homography matrix to achieve temporal consistency and also apply a \emph{PoseNet}~\cite{Kendall2015PoseNetConvolutionalNetwork} to regress translation and quaternion vectors.

\section{\acrshort{method}: Keypoint-less Calibration}\label{sec:method}

%
%
After modeling the calibration object and camera model~(\cref{sec:problem_formulation}), 
we propose the differentiable objective function~(\cref{sec:objfn}) that aims to approximate individual camera parameters given segment correspondences by iteratively minimizing the \emph{segment reprojection loss} in 2D~image space.
Finally, we introduce its direct application, the~3D~sports field registration~(\cref{sec:framework}) and required segment localization~(\cref{sec:framwork_point_prediction}). 
The main workflow is summarized in Fig.~\ref{fig:workflow}.

\subsection{Calibration Object \& Camera Model}\label{sec:problem_formulation}
%
%
Given a calibration object~(with known dimensions) that can be divided into individual labeled sub-objects of fundamental primitives~(in this paper called \emph{segments}) like \emph{points}, \emph{lines}, or \emph{point clouds}, the aim is to predict the underlying camera parameters~$\phi$ and potential lens distortion coefficients~$\psi$ that minimize its reprojection error.

\paragraph{Modeling the Calibration Object:}
%
%
Line segments are defined in the parametric form 
    $s_{\mathit{line}}=\{X_0 + \lambda X_1 | \lambda \in [0, 1]\}$ 
and point cloud segments as 
    $s_{pc}=\{X_j \in \R^3 | j=1, \hdots,|s_{pc}|\}$.
Without loss of generality, we define a labeled point segment as $s_\mathit{point}=X \in \sR^3$, resulting in the traditional \acrfull{pnp} formulation where 2D-3D point correspondences are given.
Finally, the calibration object is the composition of all individual segments per segment category~$\mathcal{C}$:
    $\sS= \bigcup\nolimits_{\mathcal{C} \in {\{\text{point, line, pc}\}}} \{ s_{\mathcal{C}}^{(1)}, s_{\mathcal{C}}^{(2)}, \hdots\}$

\paragraph{Modeling the Soccer Field:}
A soccer field is composed of lines and circle segments~(modeled as point clouds), representing all field markings, goal posts, and crossbars.
Please note that keypoint correspondences are not directly used in our approach, since all potential visible keypoints are part of line segments.
Nevertheless, we do not intend to exclude the possible explicit use of them here beforehand.
We follow the segment definitions of \citet{cioppa2022soccernet}, but modify the \emph{central circle} and split it into two parts from a heuristic in a post-processing step after semantic segmentation to induce context information.
In case of a vertically oriented \emph{middle line}, all points of the \emph{central circle} that lie on the left are assigned to a sub-segment \emph{left}, otherwise they are assigned to the sub-segment \emph{right}.

\paragraph{Modeling the Pinhole Camera:}\label{sec:cam_definition}
We use the common pinhole camera model~$\mP=\mK\mR\left[\mI | -\vt\right] \in\R^{3\times4}$ parameterized with the intrinsics \(\mK\in\R^{3\times 3}\), which define the transformation from camera coordinates to image coordinates, and extrinsics~\(\left[\mR\in\R^{3\times 3}, \vt\in\R^{3}\right]\), defining the camera pose transformation from the scene coordinates to the camera coordinates.
We assume square pixels, zero skew and set the principal point to the center of the image. 
Instead of predicting the focal length directly, i.e., the only unknown variable in $\mK$, we predict the \acrfull{fov} and transform the image coordinates to \acrfull{ndc} for numerical stability~(\cref{apx:camera}).
Following Euler's angles convention, the rotation matrix $\mR= \mR_z(roll)\mR_x(tilt)\mR_z(pan)$ is the composition of individual rotation matrices, encoding the \emph{pan}, \emph{tilt}, and \emph{roll} angles~(in radians) of the camera base according to a defined reference axis system.
Intrinsics and extrinsics are thus only parameterized by $\phi=(\acrshort{fov}, \vt, pan, tilt, roll)$, and assume that $\pi_{\phi}: X \mapsto x$  projects any scene coordinate $X\in\R^3$ to its respective image point $x\in\R^2$.

\emph{Relation to the Homography Matrix}:
If $X_z=0.0$ then $\mP^{3\times[1,2,4]}=\mK\,\mR^{3\times[1,2]}[\mI|-\vt]=\mH\in\sR^{3\times 3}$ is the respective homography matrix only able to map all points lying on one plane.
\cref{apx:hdecomp} describes how to approximate~$\phi$~given a predicted~$\hat \mH$ only.

\emph{Lens Distortion}:
As we do not want to restrict to a specific lens distortion model~$\psi$~(e.g.,~\citet{brown1966decentering}), we define $\mathtt{distort}_\psi(x)$ that distorts a pixel $x$ and $\mathtt{undistort}$ for its inverse function.
In case lens distortion coefficients are not known \emph{a priori}, we assume that  $\mathtt{undistort}$ is differentiable which enables the possibility to jointly optimize~$\psi$ and $\phi$.

\subsection{Segment Reprojection Loss}\label{sec:objfn}
\acrfull{pnp} refers to the problem of estimating the camera pose~(extrinsics) from a calibrated camera~$\mK$ given $n$ 2D-3D point correspondences.
Geometric solvers for \acrshort{pnp} or \acrshort{pnpf}, that also estimate the focal length, approximate the projection matrix~$\mP$ through the geometric or algebraic reprojection error for~$argmin_\mP~d(\vx, \pi_\mP(\mathbf{X}))$ where $d(x, \hat x)$ is the Euclidean distance between two pixels.
%
%
However, accurate correspondences are assumed to be known, the focal length in $\mK$ needs to be estimated, and there are some further requirements (e.g., minimum number of points, number of points that are allowed to be on one plane, etc.) need to be considered~\cite{Harltey2006Multipleviewgeometry}.

%
%
Instead, we aim to learn the underlying camera parameters $\phi$~(and potential lens distortion coefficients $\psi$) by minimizing the Euclidean distance between all reprojected segments and respective annotated~(or~predicted) pixels~(see \cref{sec:framwork_point_prediction} for segment localization).
Our \emph{segment reprojection loss} is based on the Euclidean distance between annotated pixels with respective segment label and reprojected segments of the calibration object.

%
%
Let us consider a sample-dependent number of pixel annotations $\vx^{(c)}\in\R^{?\times 2}$ for each~(visible) segment label~$c\in\sS$.
For a respective line segment~$s_{line}^{(c)}$, the perpendicular distance to its respective reprojected line~$\hat s_{\mathit{line}}^{(c)} = \{\pi_\phi(X_0^{(c)}) + \lambda \pi_\phi(X_1^{(c)}) | \lambda \in \R\}$~can be computed for each~$p\in\vx^{(c)}$:
\begin{equation}\label{eq:point_line_distance}
    d(p, \hat s_{\mathit{line}}) = \frac{|det((\pi_\phi(X_1) - \pi_\phi(X_0)); (\pi_\phi(X_0) - p))|}{|\pi_\phi(X_1) - \pi_\phi(X_0)|}
\end{equation} and hence describes the point-line distance.
The distance between a pixel $p^{c}\in\sR^2$ and its corresponding reprojected point cloud $\hat s_{pc}^{c}=\{\pi_\phi(X_j)|j=1,\hdots,|s_{pc}^{c}|\}$ is the minimum Euclidean distance for each $p\in\vx^{(c)}$. 
The $\mathtt{mean}$ distance over all annotated points $\vx$ is taken to aggregate one segment~$c$. 
Finally, the \emph{segment reprojection loss function} needs to be minimized where each segment contributes equally:
\begin{equation}\label{eq:loss_total}
    \gL:= \underset{\phi,~(\psi)}{argmin}\quad  \frac{1}{|\sS|} \sum_{c \in \sS} d_{\mathtt{mean}}(\mathtt{undistort}_\psi(\vx^{(c)}), \pi_\phi(s^{(c)}))
\end{equation}
Please note that $\pi$ in \cref{eq:loss_total} represents the reprojection of an arbitrary segment $\hat s=\pi_\phi(s)$ to the image to simplify the notation.
Depending on the segment type, point$\leftrightarrow$point, point$\leftrightarrow$line, or point$\leftrightarrow$point-cloud distances are computed.
Without lens distortion correction, $\mathtt{undistort}$ can be considered as identity function.

\paragraph{Implementation details:}
All computations~(image projection and distance calculation) can be performed on tensor operations, which allows for more efficient computation and parallelization. 
The input dimension of annotated or predicted pixels for each segment category~$\mathcal{C}$~(e.g., lines)~is $\hat{\tens x}_\mathcal{C}\in\sR^{T \times S_\mathcal{C} \times N_\mathcal{C} \times 2}$, where~$N_\mathcal{C}$ represents the number of selected pixels~($N_\text{keypoint}=1$),~$S_\mathcal{C}$ is the number of segments for the specific segment category, and~$T$ is an optional batch or temporal dimension.
However, we need to pad the input if the number of provided pixels per segment differ, and remember its binary \emph{padding mask}~$\tens m_\mathcal{C}\in\{0, 1\}^{T \times S_\mathcal{C} \times N_\mathcal{C}}$.
To reproject the 3D object, all points are projected from the following input dimension per segment type $\tens X_\text{line}\in\sR^{T \times S_\text{line} \times 2 \times 3}$, $\tens X_\text{pc}\in\sR^{T \times S_\text{pc} \times N_{\mathit{pc}}^* \times 3}$, and $\tens X_\text{keypoint}\in\sR^{T \times S_\text{keypoint} \times 1 \times 3}$ where~$N_{\mathit{pc}}^*$ is the number of sampled~3D~points for each point cloud.
After distance calculation for each segment type, the distance of padded input pixels are set to zero according to the \emph{padding mask} of each segment category~$\tens m_\mathcal{C}$, implying that the distance of non-visible segments is also set to zero.
Aggregating the $S$ and $N$ dimension via $\mathtt{sum}$ and dividing by the number of actually provided pixels of the input is equivalent to \cref{eq:loss_total}, where each segment contributes equally.

\subsection{Gradient-based Iterative Optimization}\label{sec:framework} 
Given human annotations or a model~(\cref{sec:framwork_point_prediction}) that predicts pixel positions with corresponding segment label, one way is to directly optimize the proposed objective function~(\cref{eq:loss_total}) via gradient descent.

\paragraph{Initialization:}
We do not further encode the camera parameters nor modify the modeled pinhole camera~(\cref{sec:cam_definition}), but rather aim to predict all unknown variables~$\phi=\{\acrshort{fov}, pan, tilt, roll, \vt \}$ in a direct manner.
However, it is beneficial to initialize an optimizer with an appropriate set of parameters.
We introduce some prior information restricting possible camera ranges.
Raw camera parameters are standardized to a zero mean and provided standard deviation.
For uniformly distributed camera ranges~$\mathcal{U}(a, b)$, we transform to a normal distribution $\mathcal{N}(\mu, \sigma)$, so that $\sigma$ covers the~$~95\%$~confidence interval, given $\mu=a+(b-a)/2$ and finally initialize with zeros.
Roughly speaking, this initialization corresponds to the mean image, e.g., a central view of the calibration object.

\paragraph{Multiple Initialization:}\label{method:multiple-initialization}
In case there is a large variance for some parameter, for instance, the camera location, it is reasonable to provide multiple sets of camera distributions.
Suppose this information is \emph{a priori}, for instance, the main broadcast camera. In that case, a user can select the correct set, or this information is known from shot boundary and shot type classification~(later denoted as $\mathtt{stacked}$).
Otherwise, we propose to run the optimization with multiple candidates and the best result is taken automatically by selecting the one with minimum loss~($\mathtt{argmin}$) according to~\cref{eq:loss_total}.

\paragraph{Self-Verification:}\label{method:self-verification}
Self-verification aims to identify all images in which the model is unable to calibrate or estimate the homography.
While other approaches use the mean point reprojection error~(e.g., \cite{Nie2021RobustEfficientFramework}) or verify geometrical constraints~\cite{Citraro2020Realtimecamera}, 
we can directly reject all samples whose loss~(\cref{sec:objfn}) is below a threshold~$\tau\in\sR^+$.
This user-defined threshold controls the trade-off between accuracy and completeness ratio and can be found empirically, e.g., by taking the best global result on a target metric for a dataset.
This procedure might be necessary for invalid input images,~e.g., out of camera distribution, erroneous semantic segmentation, or internal errors during optimization such as local minima.

\subsection{Segment Localization \& Point Selection}\label{sec:framwork_point_prediction}
The output of any model for the segment localization which provides pixel annotations for each visible segment given a raw input image can serve as input for the calibration module as well as manual annotations. 
We use the \emph{DeepLabV3 ResNet}~\cite{chen2017rethinking} (Residual Networks) to perform instance segmentation for each visible line or circle segment and do not directly predict appropriate pixels per segment.
Pixel selection is then a post-processing step, aiming to select, for instance, at least two points for a line segment with maximum distance, best representing a line where we follow a non-differentiable implementation~\cite{snchallenge}.
Ideal lines are sufficiently represented by two points, however, we have noticed more stable gradients if more than two points are selected. 
Further, we want to allow potential for lens distortion correction based on the extracted points which may show a curved polyline.

\section{Experiments}\label{sec:exp}

The experimental setup including the baselines, metrics, datasets, and hyperparameters is introduced in \cref{exp:setup}.
The results and comparisons to the state of the art are presented in \cref{exp:results}.
We conduct ablation studies for the proposed (1)~segment localization, (2)~self-verification, (3)~multiple camera initialization, and (4)~lens distortion~(\cref{exp:ablation}), while limitations are discussed in~\cref{exp:limitations}.

\subsection{Experimental Setup}\label{exp:setup}

\subsubsection{Baselines \& State of the Art}
Team sports such as soccer are played on an approximately planar field, hence many approaches assume a 2D area and use homography estimation~\cite{chu2022sports, Shi2022SelfSupervisedShape, Sha2020EndEndCamera, Nie2021RobustEfficientFramework} to map all segments lying on this plane.
To additionally estimate the camera pose and focal length, a reasonable approach is therefore the homography decomposition~(see \cref{apx:hdecomp} for details) denoted as~HDecomp.

Since in TV broadcasts of games like soccer or basketball, individual field segments are primarily visible, rather than keypoints, a suitable baseline is homography estimation via \acrshort{dlt} from line segments~\cite{snchallenge}. 
Further, we compare to \citet{chen2019sports} for homography estimation. As their retrieval and refinement module solely relies on synthetic data, we can test different variants for camera parameter distributions during training~\cite{theiner2022extraction}.
For a fair comparison, we neglect the impact of the original segment localization by using ground-truth masks generated from the SN-Calib annotations or use the predicted masks from our segmentation model.
As a second approach, we apply the official implementation from \citet{Jiang2020OptimizingLearnedErrors}.
\citet{Jiang2020OptimizingLearnedErrors} and other recent approaches~\cite{chu2022sports, Nie2021RobustEfficientFramework, Sha2020EndEndCamera} rely on annotated homography matrices for training.

\subsubsection{Datasets}

\begin{table}[bt]
\setlength{\tabcolsep}{2.5pt}
\setlength\extrarowheight{-0.5pt}
\begin{center}
\caption{Dataset comparison regarding camera type distribution, number of images, and resolution. The values labeled with $^*$ are approximated from 100 images since our calibration module does not require training data.
}
\label{tab:dataset_stats}
\fontsize{8}{8}\selectfont
\input{tables/datasets_stats}
\end{center}
\end{table}

\noindent\textbf{SN-Calib dataset:}
The SoccerNetV3-Calibration dataset~\cite{cioppa2022soccernet} consists of $\num{20028}$ images taken from the SoccerNet~\cite{deliege2021soccernet} videos~(500 matches) and covers more camera locations in addition to the main~broadcast camera.
An example setting may consist of two cameras that are placed also on the same tribune as the central broadcast camera, but are closer located to the side lines~(main camera left and right).
In addition, there are other cameras, e.g., \emph{behind the goal} and \emph{inside the goal}, or \emph{above the field}~(spider cam). 
We have manually annotated these camera locations used in this paper to get an overview. 
\cref{tab:dataset_stats} summarizes the camera type distribution and number of images per split~(train, validation, test) without stadium overlap.
Cioppa et al. provide annotation for all segments of the soccer field~\cite{cioppa2022soccernet}, i.e., lines, circle segments, and goal posts. Each visible segment has at least two annotated positions optimally representing the segment~(i.e., corner and border points) in form of a polyline.

\noindent\textbf{WC14 dataset:}
The WC14 dataset~\cite{Homayounfar2017SportsFieldLocalization} is the traditional benchmark for sports field registration in soccer and contains images from broadcast TV videos~(only central main camera without large zoom) from the FIFA World Cup 2014 and the corresponding manually annotated homography matrices.
We have additionally annotated the segments in the test split according to the guidelines in~SN-Calib~\cite{cioppa2022soccernet}.

\subsubsection{Metrics}

The quality of estimated camera parameters or homography matrices can be evaluated both at 2D image space by measuring a \emph{reprojection error}, and in world space by measuring a \emph{projection error}.

\noindent\textbf{Accuracy@threshold~\cite{snchallenge}}:
The evaluation is based on the distance of the reprojection of each soccer field segment and the corresponding annotated polyline.
Segments are reprojected from the predicted camera parameters~$\phi$~(and $\psi$) to the image from dense sampled points of the 3D model resulting in one polyline for each segment.
A polyline corresponding to a soccer field segment~$s$ is detected as a true positive~(TP), if the Euclidean distance between \textbf{every} point of the annotated polyline of segment~$\tilde{s}$ and the reprojected polyline $\pi_\phi(s)$ is less than $t$~pixels: $\forall p\in \tilde{s}: d(p, \pi_\phi(s)) < t$.
If the distance of one annotated point to its corresponding projected polyline is greater than $t$ pixels, this segment is counted as a false positive (FP), along with the projected polyline that does not appear in the annotations. 
Segments that are only present in the annotations are counted as false negatives~(FN).
Finally, the accuracy for a threshold of $t\in\{5, 10, 20\}$ pixels is given by: $AC@t = TP/(TP+FN+FP)$.
If the camera calibration or the homography estimation may fail for some images, the \textbf{Completeness Ratio}~(CR)~measures the number of provided parameters divided by the number of images of the dataset.
\noindent\textbf{Compound Score}~(CS): To summarize the above four scores, they are weighted as follows~\cite{snchallenge}:
\begin{equation}
    CS := (1 - e^{-4 CR}) \underset{t\in[5, 10, 20],w\in[0.5, 0.35, 0.15]}{(\sum w AC@t)}   
\end{equation}

\noindent\textbf{\acrfull{iou}~\cite{Homayounfar2017SportsFieldLocalization}:}
The accuracy for homography estimation for sports fields is traditionally evaluated on the $IoU_\mathit{part}$ and $IoU_\mathit{whole}$ metrics that measure the projection error.
They calculate the binary $IoU$ of the projected templates from predicted homography and a ground-truth homography in world~(top view / bird view) space for the visible area~(\emph{part}) and the full~(\emph{whole}) area of the sports field, respectively.
Due to the absence of ground-truth information like camera parameters, the evaluation can only be performed given \emph{annotated} homography matrices~\cite{Homayounfar2017SportsFieldLocalization} that are obtained from the visible sports field in the image~(e.g., via DLT). 
Hence, projection correctness can be guaranteed only for the visible area and we prefer the usage of $IoU_\mathit{part}$ similar to \citet{Nie2021RobustEfficientFramework}.

\subsubsection{Hyperparameters} 
\textbf{Optimization:}
We use AdamW~\cite{loshchilov2017decoupled} with a learning rate of $0.05$ and weight decay of $0.01$ to optimize the camera parameters~$\phi$ for $\num{2000}$ steps using the one-cycle learning rate scheduling~\cite{smith2019super} with $pct_{\mathit{start}}=0.5$. 
These parameters were found on the SN-Calib-valid split through a visual exploration of qualitative examples.
\textbf{Calibration Object \& Camera Parameter Distribution:}
Furthermore, we set the number of sampled points for each point cloud to $N_{\mathit{pc}}^*=128$~($0.45\,m$~point density for the central circle).
We use a very coarse camera distribution~(see \cref{apx:cam_distr}) of the main camera center and apply it to all datasets. 
\textbf{Segment Localization:}~The training data are derived from the provided annotations of the SN-Calib-train dataset. For training details we refer to \cref{apx:segment_localization}. 
Please recall that the expected dimension for each segment category~$\mathcal{C}$ is $\hat{\tens x}_\mathcal{C}\in\sR^{T \times S_\mathcal{C} \times N_\mathcal{C} \times 2}$.
We set {\small $|N_{\mathit{line}}|=4$} and {\small $|N_{\mathit{pc}}|=8$} following initial considerations~(\cref{sec:framwork_point_prediction}) which is in general in line with the number of annotated points per segment in SN-Calib.

\textbf{Self-Verification:}~We set the parameter~{\small$\tau=0.019$}~(\cref{method:self-verification}) globally for all experiments  
based on the maximum \emph{CS} on SN-Calib-valid-center using the predicted segment localization~(from {\small$\tau \in[0.013; 0.025]$} with a step size of {\small$10^{-3}$}; see \cref{fig:tau_multiple_splits} for visual verification).

\subsection{Results \& Comparison to State of the Art}\label{exp:results}

Previous approaches focus on the~(1)~main camera center and~(2)~homography estimation. Hence, we~(1)~compare on the subset of SN-Calib-test and~(2) measure both the camera calibration performance induced by the predicted camera parameters and the homography estimation.

\paragraph{Reprojection Error for Camera Calibration:}
This task represents the main task of estimating individual camera parameters~$\phi$ where the reprojection error~(\emph{AC@t}) induced by~$\phi$ is evaluated.
The results on the test splits on SN-Calib-center and WC14-test are presented in \cref{tab:main_results_testset}~(top) and \cref{tab:wc14_results}, respectively.

\emph{Pred vs. Ground Truth (GT) Segmentation}: 
If the same ground-truth segmentation is used as input, our method outperforms the best variant from \citet{chen2019sports}~($\mathcal{U}_{FoV}$+$\mathcal{U}_{xyz}$~\cite{theiner2022extraction}) and the baseline on both datasets.

\emph{Self-Verification}: The homography decomposition also contains a kind of self-verification resulting in a higher reprojection accuracy~(AC@t) but lower completeness ratio~(CR), as shown in \cref{tab:main_results_testset,tab:wc14_results}.
Hence, we can compare these approaches with our results after self-verification of \acrshort{method}. 
Superior results are achieved for all variants of segmentation and on both datasets.

\paragraph{Reprojection Error for Homography Estimation:}
To investigate whether the quality of the homography estimation or the decomposition are the reason for the results, we examine the plain performance of the homography estimation and thus exclude the impact of the homography decomposition.
We measure the same metrics, but only map all segments lying on one plane, i.e., ignore goal posts and crossbars.
The results for the estimated as well as the ground-truth homography matrices are presented in \cref{tab:main_results_testset}~(bottom) and \cref{tab:wc14_results_homography}.

\emph{Influence of the Homography Decomposition}:
Compared to the reprojection error for the calibration task, noticeably better results are achieved indicating that the decomposition introduces additional errors.
Based on the per-segment accuracy, we found that in particular a larger projection error is frequently visible for goal segments since the height information is missing but not the only reason for higher errors~(e.g., DLT Lines with and without HDecomp).

\emph{Pred vs. GT Segmentation}: 
Similar to the evaluation of camera calibration performance, superior results are achieved on both datasets with a noticeable drop when using the segment localization model instead of ground truth.
The evaluation on the WC14 dataset~(\cref{tab:wc14_results_homography}) yields better results when using segment localization from the individual approaches~\cite{chen2019sports, Jiang2020OptimizingLearnedErrors} trained on this dataset, but still the \acrshort{method} approach outperforms these variants.

\paragraph{Projection Error~(\acrshort{iou}):}
\acrshort{method} achieves very similar results compared to the reproduced approaches and other state-of-the-art approaches without performing training or fine-tuning on this dataset.
The reprojection error measured via AC@t from the annotated homographies~$\mH$ is comparable with our results~(\cref{tab:wc14_results_homography}), but not ideal, demonstrating bias on the \acrshort{iou} metrics since~$\mH$ is used to evaluate the projection error.

\begin{table}[bt]
\setlength{\tabcolsep}{2.8pt}
\setlength\extrarowheight{0.6pt}
\begin{center}
\caption{Results on SN-Calib-test-center only evaluating where the main camera center is shown~(1454 images): When evaluating the homography, all segments not lying one the plane~(goal posts and crossbars) are ignored.}
\label{tab:main_results_testset}
\fontsize{8}{8}\selectfont
\input{tables/sncalib-test-center}
\end{center}
\end{table}
\begin{table}[bt]
\setlength{\tabcolsep}{2.8pt}
\setlength\extrarowheight{0.6pt}
\begin{center}
\caption{Evaluating the reprojection error induced by the camera parameters~($\phi$)~on WC14-test dataset~(186 images).
}
\label{tab:wc14_results}
\small
\fontsize{8}{8}\selectfont
\input{tables/wc14-test}

\end{center}
\end{table}
\begin{table}[bt]
\setlength{\tabcolsep}{2.1pt}
\setlength\extrarowheight{0.6pt}
\begin{center}
\caption{Evaluating the homography estimation on WC14-test: $IoU_\text{part}$ compares the projection error~(top view) using annotated homography matrices~($\mH$~\cite{Homayounfar2017SportsFieldLocalization}). \textcolor{gray}{Grayed out}: Results taken from the respective paper. 
}
\label{tab:wc14_results_homography}
\small
\fontsize{8}{8}\selectfont

\input{tables/wc14-test-homography}
\end{center}
\end{table}

\subsection{Ablation Studies}\label{exp:ablation}

\begin{figure}[bt]
\begin{center}
  \includegraphics[width=\linewidth]{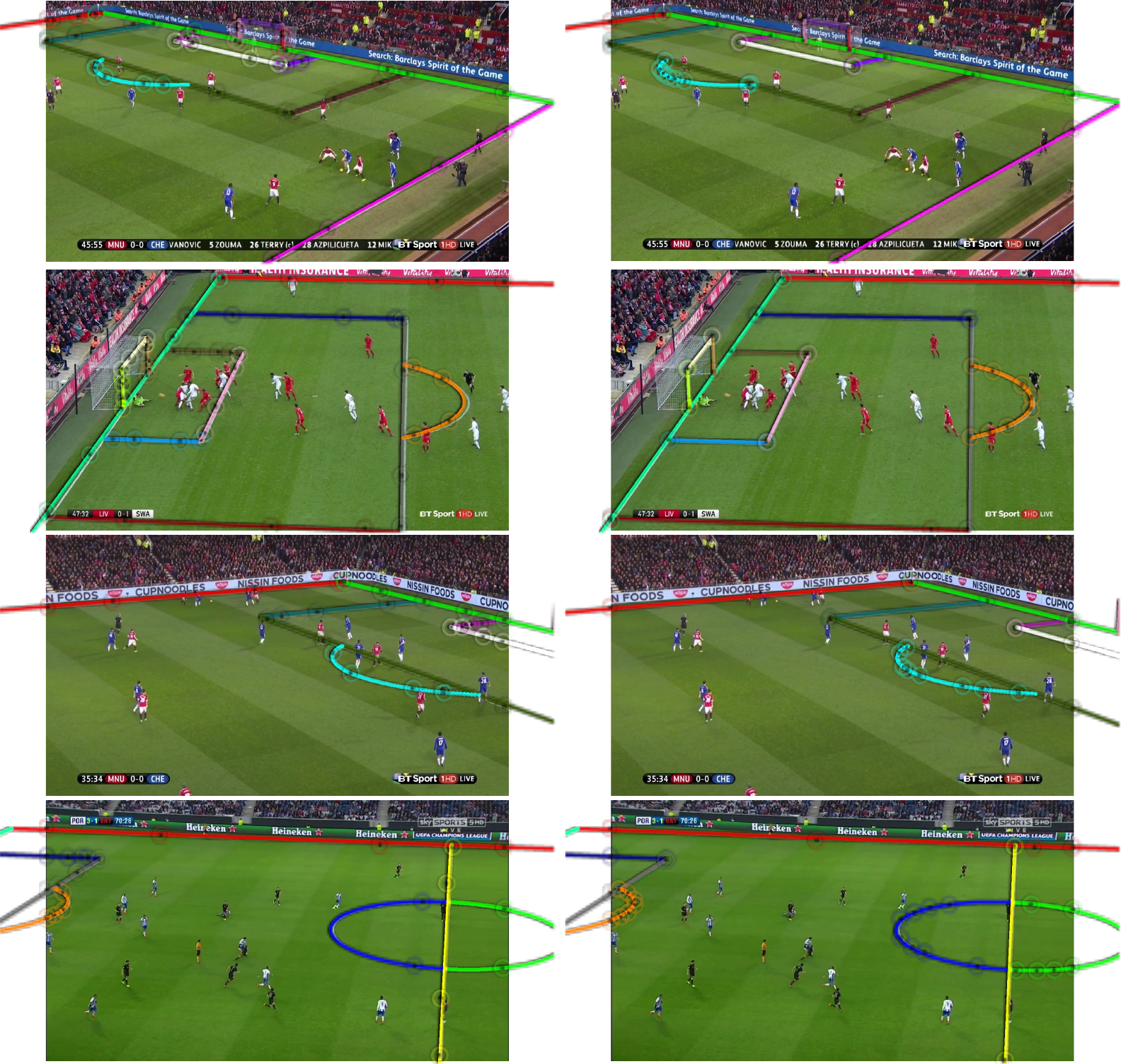}
\end{center}
  \caption{Random samples for \acrshort{method}~($\mathtt{argmin}$) on SN-Calib-test using predicted~(left) and GT (right) segments.}
\label{fig:main_results_testset_images}
\end{figure}

\paragraph{Impact of Segment Localization (Pred vs. GT):}
Because we want to find the upper limit for the performance of our method, we use the provided annotations and compare with the predicted segments from our segment localization model.  
The lower performance~(\cref{tab:main_results_testset,tab:wc14_results}) when using the predicted segments shows that the \emph{segment localization} module~(\cref{sec:framwork_point_prediction}) needs improvement despite the visually similar results for the majority of images~(\cref{fig:main_results_testset_images}).

\begin{figure}[tb]
  \centering
  \includegraphics[width=0.99\linewidth]{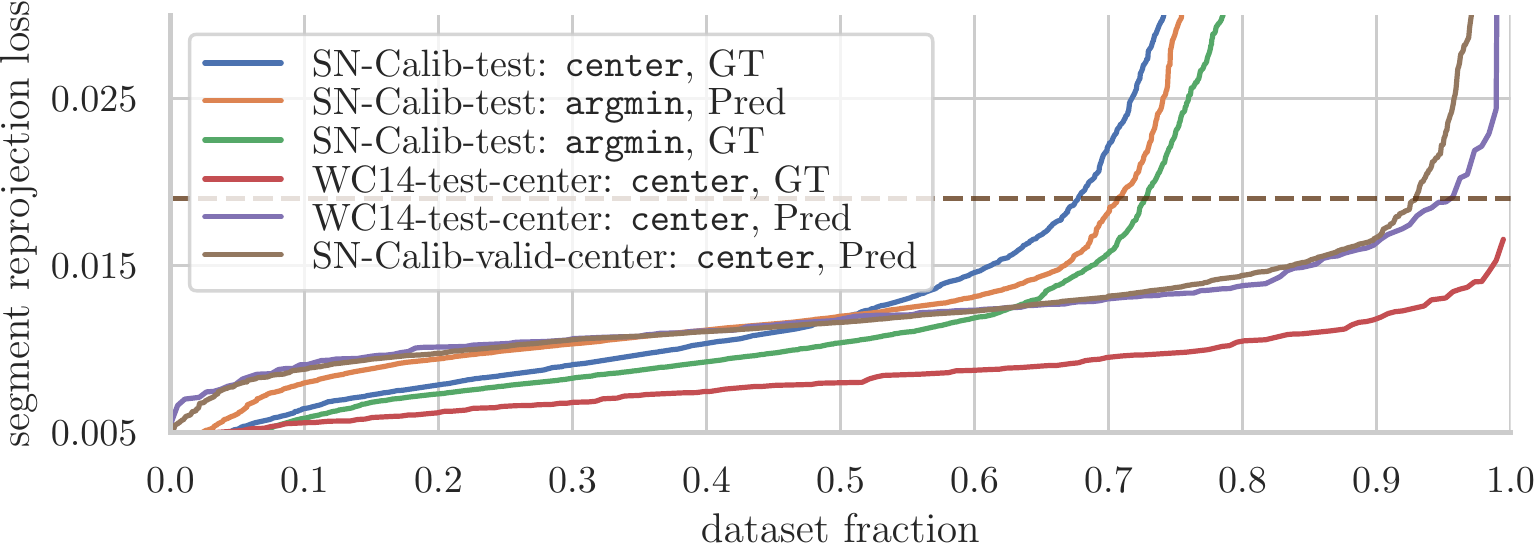}
  \caption{\emph{Segment reprojection loss} per sample for several dataset splits and configurations.} 
  \label{fig:tau_multiple_splits}
\end{figure}

\begin{figure}[tb]
  \centering
   \includegraphics[width=1.0\linewidth]{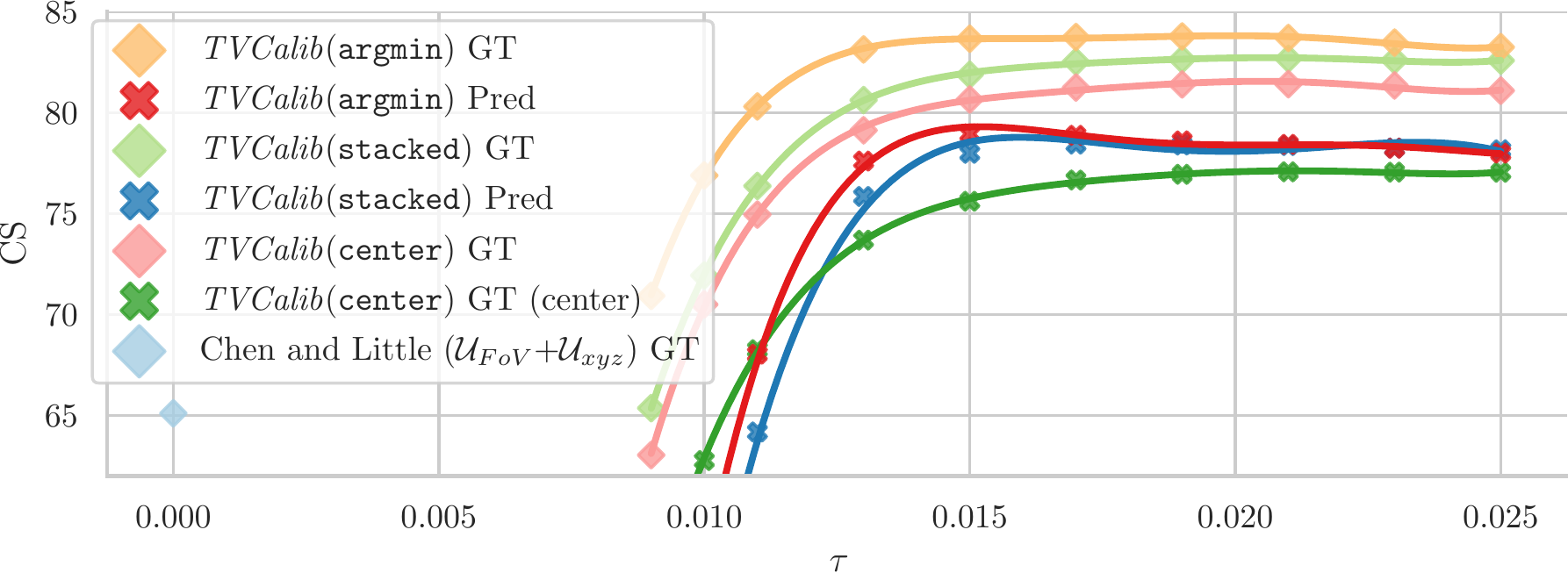}
   \caption{Aggregated results on SN-Calib-test~(all) for the calibration task:  Different variants of \acrshort{method} are shown for several self-verification thresholds~$\tau$.
   }
   \label{fig:main_results_testset}
\end{figure}

\paragraph{Choice of the Self-Verification Parameter:}\label{exp:self-verification}
Please recall that~$\tau$ is a user-defined threshold able to reject images based on the \emph{reprojection loss}.
For simplicity, we have set this value once globally based on the maximum \emph{CS} on SN-Calib-valid-center~(predicted segment localization), but the optimal value can be chosen for each dataset and configuration individually or specified manually.
This value is roughly valid across multiple datasets, camera distributions, and splits~(see \cref{fig:tau_multiple_splits}).
The projection performance is shown in \cref{fig:main_results_testset} for multiple configurations of \acrshort{method} by varying this parameter.
In general, the more~$\tau$ is restricted, the less the completeness ratio decreases, with increasing accuracy that at some point saturates.

\paragraph{Multiple Initialization:}
As our solution aims to optimize the camera parameters for multiple camera locations~($\mathtt{center, left, right}$),~(1)~the question arises whether one initialization~($\mathtt{center}$) is sufficient or multiple initialization~(one per camera location) are preferred, and~(2), if the camera position is known \emph{a priori}, one variant is to use only the respective initialization and for this experiment to stack the results~($\mathtt{stacked}$). The other variant utilizes the optimization from multiple initializations and takes the best result~($\mathtt{argmin}$).
As shown in \cref{fig:main_results_testset}, initializing from three camera positions~($\mathtt{argmin}$ and $\mathtt{stacked}$) is noticeably better than using only one initialization~($\mathtt{center}$), and selecting the best result~($\mathtt{argmin}$) is slightly better than knowing the camera type in advance~($\mathtt{stacked}$). 
Due to the iterative optimization process, the ability to start from several locations enables the chance to find better minima.

\paragraph{Lens Distortion:}
The results when camera and radial lens distortion parameters were learned jointly are presented and discussed in \cref{apx:lens_dist}. 
In summary, results can be improved at \emph{AC@5} for samples where radial lens distortion is visible.

\subsection{Limitations}\label{exp:limitations}
Despite strong results, for a small fraction of given ground-truth segment annotations, some samples are rejected.
This is mainly caused by local minima due to the nature of gradient-based iterative optimization~\cite{acuna2018insights}.
Related to the camera initialization, we have not investigated any cameras other than those on the main tribune.
The \acrshort{method} approach relies on an accurate segment localization, but no regularization term is included that allows for outliers. 
Finally, jointly learning lens distortion coefficients has not been deeply investigated.

\section{Conclusions}\label{sec:conclusions}
We have presented an effective 
solution to learn individual camera parameters from a calibration object that is modeled by~point, line, and point cloud segments. 
Furthermore, we have successfully demonstrated its direct application to 3D~sports field registration in soccer broadcast videos.
In the target task of 3D as well as for 2D~sports field registration, our method has achieved superior results compared to two state-of-the-art approaches~\cite{Jiang2020OptimizingLearnedErrors, chen2019sports} for~2D~sports field registration in terms of the image reprojection error.

Future work could investigate the integration of temporal consistency and associated speedup, the application to other sports, and finally the incorporation into a deep neural network to estimate the camera parameters in one feed-forward step or full end-to-end learning.

\section*{Acknowledgement}
{
Thanks to Wolfgang Gritz and Eric Müller-Budack for reviewing this paper, Jim Rhotert for the segmentation module and Markos Stamatakis for enriching the WC14 dataset.
This project has received funding from the German Federal Ministry of Education and Research (BMBF~--~Bundesministerium für Bildung und Forschung) under~01IS20021B.
}

\appendix

\section{Camera Model}

\subsection{Field of View at NDC and Raster}\label{apx:camera}
For numerical stability the input image or pixel are normalized to image dimensions from $[-1, 1]$~(\acrshort{ndc}) and the \acrshort{fov}~(in radian) is predicted instead of the focal length resulting in $f_x^{NDC}=\frac{1}{tan(0.5 \times \acrshort{fov})}$ and $f_y^{NDC}= a \times \frac{0.5}{tan(0.5 \times \acrshort{fov})}$ where $a$ is the original image aspect ratio. To access the true focal length, we know square pixel and thus, use $f_x^{I}=f_y^{I}=w \times \frac{0.5}{tan(0.5 \times \acrshort{fov})}$ where $w$ is the original image width in pixel.

\subsection{Camera Distribution}\label{apx:cam_distr}
The following camera parameter distribution cover a variety of stadiums over the world for the main tribune and is coarser distribution as used in \cite{chen2019sports}: 
{\small
$pan \in \mathcal{U}(-45^{\circ}, 45^{\circ})$, 
$tilt \in \mathcal{U}(45^{\circ}, 90^{\circ})$, 
$roll \in \mathcal{U}(-10^{\circ}, 10^{\circ})$, 
$aov \in \mathcal{U}(8.2^{\circ}, 90^{\circ})$, 
$t_z \in \mathcal{U}(-40\,m, -5\,m)$,
$t_y \in \mathcal{U}(40\,m, 110\,m)$,
$t_x \in \mathcal{U}(-40\,m, -5\,m)$} for main camera center, {\small $t_y \in \mathcal{U}(-36 - 16.5\,m, -36 + 16.5\,m)$} for main camera left and {\small$t_y \in \mathcal{U}(36 - 16.5\,m, 36 + 16.5\,m)$} for main camera right.

\subsection{Note on the World Reference Axis System}\label{apx:sota_alignment}
Given a \href{https://github.com/SoccerNet/sn-calibration/blob/main/doc/axis-system-resized.jpeg}{world reference coordinate system} and the definition of the pinhole camera model~($\mK\mR[\mI|-\vt]$) including the principal axis, the decomposition of $\mH$ in $\mR$ and $\vt$ and individual rotation angles must follow the concrete definition in order to derive expected values.
In case where the world axis system differ~\cite{chen2019sports, Jiang2020OptimizingLearnedErrors} the provided homography matrices can be aligned.

\paragraph{Alignment with \citet{chen2019sports}:}
Because \citet{chen2019sports} place the coordinate system differently through the sports field, the output~$\hat \mH_\text{\cite{chen2019sports}}$ of the reproduced model~(reimplemented using the \href{https://github.com/lood339/SCCvSD}{official code snippets} from the authors) is aligned to the SN-Calib axis system according to
\begin{equation*}
\hat \mH = \mR(\mT\hat \mH_\text{\cite{chen2019sports}}) 
    = \Big[\begin{smallmatrix}
1 & 0 & 0\\
0 & -1 & 0 \\
0 & 0 & 1 \\
\end{smallmatrix}\Big](\Big[\begin{smallmatrix}
1 & 0 & -105/2\\
0 & 1 & -68/2 \\
0 & 0 & 1 \\
\end{smallmatrix}\Big] \hat \mH_\text{WC14})
\end{equation*}
where first the coordinate center is moved to the middle of the sports field and only the direction of the $y$-axis is swapped.

\paragraph{Alignment with WC14~\cite{Homayounfar2017SportsFieldLocalization} homography matrices:}
The provided homography matrices from the WC14 dataset~($\tilde{H}$) are aligned to the SoccerNet coordinate system as follows:
The scene coordinate center needs to be moved to the center of the sports field and the dimensions need to be scaled from yards to meters~($y2m\approx 0.9144$):
\begin{equation*}
\hat \mH = \mS(\mT\hat \mH_\text{WC14}) 
    = \Big[\begin{smallmatrix}
y2m & 0 & 0\\
0 & y2m & 0 \\
0 & 0 & 1 \\
\end{smallmatrix}\Big](\Big[\begin{smallmatrix}
1 & 0 & -115/2\\
0 & 1 & -74/2 \\
0 & 0 & 1 \\
\end{smallmatrix}\Big] \hat \mH_\text{WC14})
\end{equation*}

\paragraph{Alignment with \citet{Jiang2020OptimizingLearnedErrors}:}

\citet{Jiang2020OptimizingLearnedErrors} use $[-0.5, 0.5]$ as sports field and image template dimensions and centered origin.
The output~$\hat \mH_{\text{\cite{Jiang2020OptimizingLearnedErrors}}}$ of their \href{https://github.com/vcg-uvic/sportsfield_release}{officially provided model} is first aligned to WC14~\cite{Homayounfar2017SportsFieldLocalization}~($\hat \mH_{\text{\cite{Jiang2020OptimizingLearnedErrors}}}^*$) which is subsequently aligned to SoccerNet by scaling~(1)~the image to the original resolution~(W, D),~(2)~scaling the template image to the used~720p resolution:

\begin{equation*}
\begin{split}
    \hat \mH = \Big[\begin{smallmatrix}
W & 0 & W/2\\
0 & H & H/2 \\
0 & 0 & 1 \\
\end{smallmatrix}\Big] \hat \mH_{\text{\cite{Jiang2020OptimizingLearnedErrors}}}^* \Big[\begin{smallmatrix}
1280 & 0 & 640\\
0 & 720 & 360 \\
0 & 0 & 1 \\
\end{smallmatrix}\Big]^{-1}
\end{split}
\end{equation*}

\section{Homography Decomposition: From Homography to Camera Parameters}\label{apx:hdecomp}
This section describes how to extract the camera position~$\vt$, orientation~(\emph{pan, tilt, roll}) and \emph{focal length} from a plane homography according to the pinhole camera model as described in~\cref{sec:cam_definition} assuming square pixel, zero skew, and a centered principal point.
In general, given a calibration matrix~$\mK$ and a homography matrix~$\mH$ that describes the mapping between two planes~(e.g., derived from point correspondences), rotation matrix~$\mR$ and translation vector~$\vt$, can be derived~\cite{Harltey2006Multipleviewgeometry}.
The procedure described below is in general in line with \cite{Citraro2020Realtimecamera, snchallenge} and we mainly follow the implementation from~\cite{snchallenge}.

\noindent\textbf{(1)~Computing the Focal Length}:
As $\mH$ already describes the relation between two planes, and the focal length is the only unknown parameter in~$\mK$, the first step is to approximate the focal length~(see Algorithm 8.2 in \citet{Harltey2006Multipleviewgeometry}) given constraints from the homography matrix and our assumptions on $\mK$.

\noindent\textbf{(2)~Computing the Rotation Matrix}:
Leveraging the relation between the approximated calibration matrix and provided homography, orientation~(first rotation matrix~$\mR$) and translation~$\vt$~(camera position) are then approximated as we know that $\mH\overset{X_z=0}{=}\mP^{3\times[1,2,4]}=\mK\,R^{3\times[1,2]}[\mI|-\vt]$. 

Since~$\mK$ is already given, $\mK^{-1}\mH$ yields individual column vectors~$[\vr_1', \vr_2', -\vt']$ encoding rotation and translation. 
After normalizing $\vr_1', \vr_2'$ to unit length, the third column $\vr'_3$ of the rotation matrix~$\mR'=[\vr_1, \vr_2, \vr_3]$ can be approximated from $\vr_1\times \vr_2$, 
since we expect orthogonality for~$\mR$~(constructed from per axis rotations, i.e., $\mR_z(roll)\mR_x(tilt)\mR_z(pan)$).
Singular value decomposition is applied~$\mU\mS\mV^T=\mR'$ and since one property is that~$\mU, \mV$ are real orthogonal matrices, the estimated rotation matrix is $\mR=\mU\mV^T$.

\noindent\textbf{(3)~Computing the Camera Position}:
The translation vector is finally derived from $\vt= -\mR^T (\vt' * \sqrt{|\vr'_1|\times|\vr'_2|})$.

\noindent\textbf{(4)~Refining $\mR$ and $\vt$}:
Once $\mK$, $\mR$, and $\vt$ are roughly approximated, the camera pose can be refined given reprojected keypoints from $\mH$~(2D-3D point correspondences) via non-linear least-squares minimization~(Levenberg-Marquardt refinement, see \href{https://docs.opencv.org/4.6.0/d9/d0c/group__calib3d.html#ga650ba4d286a96d992f82c3e6dfa525fa}{\small$\mathtt{cv2.solvePnPRefineLM()}$})
As the Levenberg-Maquardt algorithm is not able to handle large refinements, a point is not considered if its reprojection error between initial estimation and homography is larger than $\zeta=100$ pixels~\cite{snchallenge}.

In contrast to \citet{snchallenge}, to provide a reasonable set of keypoint correspondences, a keypoint is only considered if a point of the homography is visible in the image with a tolerance of $0.1\times$ image width and height, respectively. 
The tolerance is motivated by a simple example: Assume the keypoint in the middle of the central circle which is close outside the visible image. It is a valuable information despite it is not visible.
In case the number of point correspondences is smaller than three, the refinement algorithm cannot be performed. 
We reject the entire sample and do not return the initial estimation as the difference between the decomposition and the original estimated homography is too large.

\noindent\textbf{(5)~Accessing Individual Rotation Angles}:
As $\mR$ is composed of individual per axis rotations representing pan, tilt, and roll of the camera of known order~(i.e., a known scene coordinate system) and given principal axis, individual rotation angles can be extracted by solving~$\mR= \mR_z(roll)\mR_x(tilt)\mR_z(pan)$ for pan, tilt, and roll angles.
However, as there are two solutions, we exploit world knowledge and take the solution where the roll parameter is minimal~\cite{snchallenge}.

\section{Segment Localization}\label{apx:segment_localization}
We use the \emph{DeepLabv3} ResNet-101~\cite{chen2017rethinking} architecture to perform instance segmentation on all sports field segments.
To train this model, we use the training data from the SN-Calib train split and validate on the respective validation split while keeping the model with the lowest loss on validation. During training, images are resized to a height of 256 pixels.
Following the \href{https://github.com/pytorch/vision/tree/main/references/segmentation}{vanilla training script} and suggested parameters, we train for max. 30 epochs using a batch size of 8, \acrshort{sgd}~(momentum: $0.9$, weight decay: $1^{-4}$), learning rate of $0.01$, initialized with \emph{ImageNet1k} weights, and \emph{auxiliar loss}.

\section{Radial Lens Distortion Correction}\label{apx:lens_dist}

\begin{table}[tb]
\setlength{\tabcolsep}{2.5pt}
\setlength\extrarowheight{2.0pt}
\begin{center}
\caption{Ablation study for radial lens distortion correction~(LD)}
\label{tab:lens_dist_results}
\fontsize{8}{8}\selectfont
\input{tables/radial_lens_distortion}
\end{center}
\end{table}

As only radial lens distortion seems to be present for some samples, optimization of radial lens distortion coefficients~$\psi=\{k_1, k_2\}$ may also be performed where we follow \href{https://kornia.readthedocs.io/en/latest/_modules/kornia/geometry/calibration/distort.html}{\emph{kornia}}'s implementation of lens distortion models.

To first focus on learning the camera parameters~$\phi$, lens distortion coefficients~$\psi$ are optimized with its own optimizer~(also \emph{AdamW} but with a learning rate of $1e^{-3}$) and one-cycle learning rate scheduling~($pct_{start}=0.33$).

\begin{figure}[tb]
  \centering
   \includegraphics[width=\linewidth]{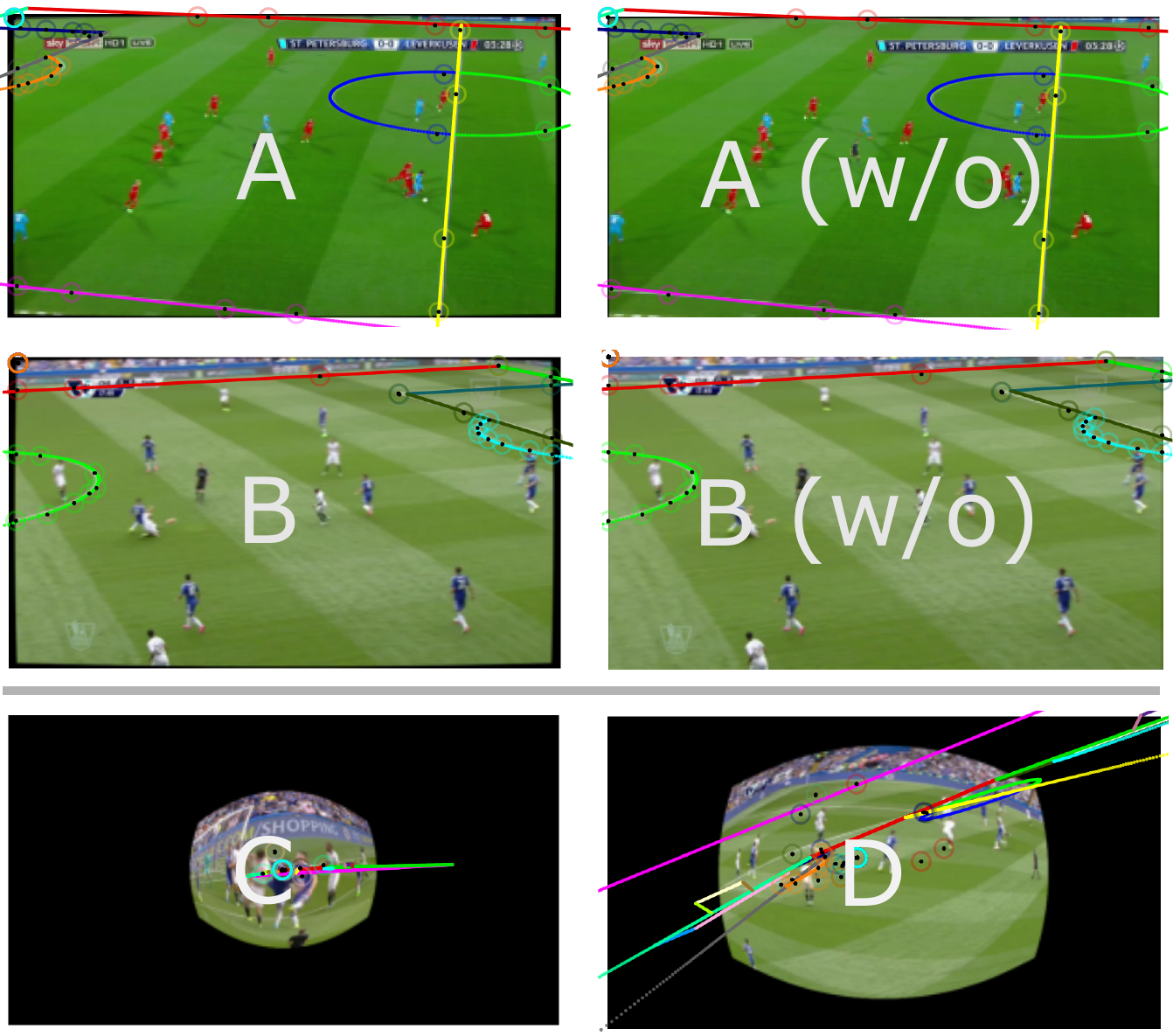}
   \caption{A, B: Samples where radial lens distortion is present~(left: with correction, right: without correction for comparison); C, D: samples with lower \acrshort{fov} result in trivial local minima when jointly optimizing distortion coefficients and camera parameters.}
   \label{fig:lens_dist_examples}
\end{figure}

We have observed this process works for many samples where radial lens distortion is present~(\cref{tab:lens_dist_results}, \cref{fig:lens_dist_examples}~A and B) with significantly better results on WC14, but noticed an issue on the SN-Calib dataset and specific samples~(e.g., \cref{fig:lens_dist_examples}~C and D, mainly images with a low \acrshort{fov}). 
WC14 is not affected as usually a larger \acrshort{fov} is shown.
Selected points are transformed via $\mathtt{undistort}$~according to \cref{eq:loss_total} and in some cases the points are distorted too much~(towards the principal point) or the \acrshort{fov} explodes, resulting in local minima.

Transforming the reprojected points~($\mathtt{distort}$) instead of the selected points is reasonable~(i.e., $d(\vx^y, \mathtt{distort}_\psi \pi_\phi(s^y))$), but the distance calculation for ideal lines is very effective and needs to be adjusted otherwise.
We continue to investigate this issue to find a practical solution. 
Results indicate that the performance can be increased for samples where radial lens distortion is present.

\newpage
{\small
\setlength{\bibsep}{0pt} 
\bibliographystyle{plainnat}
\bibliography{egbib}
}

\end{document}

%% file: math_includes.tex
\usepackage{makecell}

\usepackage{amsmath,amsfonts,bm}

\def\1{\bm{1}}

\def\vr{{\bm{r}}}

\def\vt{{\bm{t}}}

\def\vx{{\bm{x}}}

\def\mH{{\bm{H}}}
\def\mI{{\bm{I}}}

\def\mK{{\bm{K}}}

\def\mP{{\bm{P}}}

\def\mR{{\bm{R}}}
\def\mS{{\bm{S}}}
\def\mT{{\bm{T}}}
\def\mU{{\bm{U}}}
\def\mV{{\bm{V}}}

\DeclareMathAlphabet{\mathsfit}{\encodingdefault}{\sfdefault}{m}{sl}
\SetMathAlphabet{\mathsfit}{bold}{\encodingdefault}{\sfdefault}{bx}{n}
\newcommand{\tens}[1]{\bm{\mathsfit{#1}}}

\def\gL{{\mathcal{L}}}

\def\sR{{\mathbb{R}}}
\def\sS{{\mathbb{S}}}

\newcommand{\R}{\mathbb{R}}



%% file: tables/datasets_stats.tex
\begin{tabularx}{\linewidth}{Xr|rr|rrrr}
\toprule
\multirow{2}{*}{\textbf{Dataset}} & \multirow{2}{*}{\textbf{Split}} & \multirow{2}{*}{\textbf{Images}} & \multirow{2}{*}{\textbf{Reso.}} &\multicolumn{4}{c}{\textbf{Camera Type Distr.} [\%]} \\
& &   &  & \multicolumn{1}{r}{Center} & \multicolumn{1}{r}{Left} & \multicolumn{1}{r}{Right} & \multicolumn{1}{r}{Other} \\ \midrule \midrule
\multirow{3}{*}{SN-Calib}   & train & 14513 & 540p       & $^*$48.0 & $^*$14.0 & $^*$15.0 & $^*$23.0 \\
                            & valid & 2796  & 540p       & 52.7 & 10.2 & 9.5 & 27.7 \\
                            & test & 2719  & 540p        & 53.5 & 8.5 & 9.5  & 28.5 \\ \midrule
\multirow{2}{*}{WC14} & train/valid & 209 & 720p      & 100. & 0.0 & 0.0 & 0.0 \\
 & test & 186 & 720p             & 100. & 0.0 & 0.0 & 0.0 \\
\bottomrule
\end{tabularx}

%% file: tables/sncalib-test-center.tex

\begin{tabularx}{\linewidth}{Xr|rrrr|r}
\toprule

& & \multicolumn{3}{c}{\textbf{AC@} [\%]} & &   \\
\textbf{Calibration} & \textbf{Seg.} & \textbf{5} & \textbf{10} & \textbf{20} & \textbf{CR} & \textbf{CS} \\
\midrule  \midrule
\multicolumn{7}{c}{\textbf{Evaluating the Camera Calibration~($\Hat\phi$)}} \\ 
\midrule

                       \rowcolor{OliveGreen!10} \acrshort{method}($\tau$) & GT & 68.7 & 88.0 & 96.1 &  92.8 & 76.9 \\
                               \rowcolor{OliveGreen!10} \acrshort{method} & GT & 65.3 & 84.2 & 92.6 & 100.0 & 75.5 \\
HDecomp + \cite{chen2019sports} ($\mathcal{U}_{FoV}$+$\mathcal{U}_{xyz}$) & GT & 53.7 & 77.5 & 88.4 &  80.3 & 65.1 \\
                                                      HDecomp + DLT Lines & GT & 48.1 & 68.5 & 84.6 &  79.8 & 60.2 \\

\midrule  
                       \rowcolor{OliveGreen!10} \acrshort{method}($\tau$) & Pred & 57.6 & 81.7 & 93.2 &  93.7 & 72.6 \\
                               \rowcolor{OliveGreen!10} \acrshort{method} & Pred & 54.8 & 78.5 & 90.4 & 100.0 & 71.4 \\
                                                      HDecomp + DLT Lines & Pred & 40.6 & 63.2 & 80.4 &  79.6 & 55.9 \\
HDecomp + \cite{chen2019sports} ($\mathcal{U}_{FoV}$+$\mathcal{U}_{xyz}$) & Pred & 34.4 & 64.6 & 81.3 &  66.6 & 52.0 \\

\midrule
\multicolumn{7}{c}{\textbf{Evaluating the Homography Estimation $\hat\mH$}} \\  
\midrule
                       \rowcolor{OliveGreen!10} \acrshort{method}($\tau$) & GT & 65.0 & 85.4 & 95.6 &  92.8 & 75.5 \\
                               \rowcolor{OliveGreen!10} \acrshort{method} & GT & 61.7 & 81.6 & 92.0 & 100.0 & 73.9 \\
          \cite{chen2019sports} ($\mathcal{U}_{FoV}$+$\mathcal{U}_{xyz}$) & GT & 57.3 & 76.0 & 83.7 & 100.0 & 68.0 \\
HDecomp + \cite{chen2019sports} ($\mathcal{U}_{FoV}$+$\mathcal{U}_{xyz}$) & GT & 61.1 & 81.2 & 89.4 &  80.3 & 67.5 \\
                                                                DLT Lines & GT & 54.7 & 69.9 & 81.6 &  97.6 & 64.4 \\
                                                      HDecomp + DLT Lines & GT & 56.5 & 74.3 & 86.3 &  79.8 & 63.6 \\

\midrule
                       \rowcolor{OliveGreen!10} \acrshort{method}($\tau$) & Pred & 54.6 & 78.3 & 92.4 &  93.7 & 70.8 \\
                               \rowcolor{OliveGreen!10} \acrshort{method} & Pred & 51.9 & 75.2 & 89.4 & 100.0 & 69.5 \\
                                                                DLT Lines & Pred & 46.9 & 66.5 & 79.3 &  97.9 & 61.3 \\
                                                      HDecomp + DLT Lines & Pred & 46.5 & 68.5 & 83.0 &  79.6 & 59.2 \\
          \cite{chen2019sports} ($\mathcal{U}_{FoV}$+$\mathcal{U}_{xyz}$) & Pred & 32.9 & 59.0 & 72.5 & 100.0 & 54.6 \\
HDecomp + \cite{chen2019sports} ($\mathcal{U}_{FoV}$+$\mathcal{U}_{xyz}$) & Pred & 40.1 & 68.3 & 82.3 &  66.6 & 54.0 \\

\bottomrule
\end{tabularx}

%% file: tables/wc14-test.tex



\begin{tabularx}{\linewidth}{Xr|rrrr|r}
\toprule
& & \multicolumn{3}{c}{\textbf{AC@} [\%]} & &   \\
\textbf{Calibration} & \textbf{Seg.} & \textbf{5} & \textbf{10} & \textbf{20} & \textbf{CR} & \textbf{CS} \\
\midrule  \midrule

                   \rowcolor{OliveGreen!10} \acrshort{method} &           GT &  64.4 &   86.7 &   96.0 &         100.0 & 86.4 \\
                              HDecomp + \cite{chen2019sports} &           GT &  52.8 &   78.8 &   91.3 &          90.9 & 79.0 \\
HDecomp + $\mH$~\cite{Homayounfar2017SportsFieldLocalization} &       \xmark &  48.1 &   78.9 &   91.5 &          90.9 & 78.4 \\
                                          HDecomp + DLT Lines &           GT &  32.0 &   54.0 &   73.1 &          73.7 & 57.1 \\

\midrule

                    \rowcolor{OliveGreen!10} \acrshort{method} &                                    Pred &  39.9 &   71.9 &   90.5 &         100.0 & 75.0 \\
                   HDecomp + \cite{chen2019sports} $\zeta$=1k) &                   \cite{chen2019sports} &  29.0 &   59.8 &   79.0 &         100.0 & 63.6 \\
HDecomp + \cite{Jiang2020OptimizingLearnedErrors}~($\zeta$=1k) & \cite{Jiang2020OptimizingLearnedErrors} &  32.4 &   58.5 &   75.3 &          99.5 & 61.8 \\

\midrule

            \rowcolor{OliveGreen!10} \acrshort{method}($\tau$) &                                    Pred &  41.3 &   73.6 &   91.4 &          95.7 & 76.0 \\
                               HDecomp + \cite{chen2019sports} &                   \cite{chen2019sports} &  32.7 &   67.3 &   87.3 &          81.7 & 69.4 \\
             HDecomp + \cite{Jiang2020OptimizingLearnedErrors} & \cite{Jiang2020OptimizingLearnedErrors} &  36.9 &   66.4 &   83.9 &          84.9 & 68.4 \\
                               HDecomp + \cite{chen2019sports} &                                    Pred &  28.1 &   60.6 &   80.8 &          78.5 & 63.0 \\
                                           HDecomp + DLT Lines &                                    Pred &  26.9 &   53.3 &   72.7 &          74.2 & 56.0 \\

\bottomrule
\end{tabularx}

%% file: tables/wc14-test-homography.tex
\begin{tabularx}{\linewidth}{Xr|rrrr|r|rr}
\toprule
 \multirow{2}{*}{\textbf{Approach}} & \multirow{2}{*}{\textbf{Seg.}} & \multicolumn{3}{c}{\textbf{AC@} [\%]} & \multirow{2}{*}{\textbf{CR}} & \multirow{2}{*}{\textbf{CS}} & \multicolumn{2}{c}{$IoU_{\text{part}}$} \\
 &  & \multicolumn{1}{c}{\textbf{5}} & \textbf{10} &  \textbf{20} & &  & mean & med. \\ \midrule \midrule
%

           \rowcolor{OliveGreen!10} \acrshort{method}($\tau$) &           GT &  62.7 &   84.9 &   95.5 &         100.0 & 85.3 \\
                              HDecomp + \cite{chen2019sports} &           GT &  56.1 &   80.6 &   91.1 &          90.9 & 80.0 \\
HDecomp + $\mH$~\cite{Homayounfar2017SportsFieldLocalization} &       \xmark &  50.6 &   79.4 &   91.1 &          90.9 & 78.8 \\
HDecomp + DLT Lines &           GT &  35.8 &   57.6 &   74.2 &          73.7 & 59.4 \\

\midrule

                   \rowcolor{OliveGreen!10} \acrshort{method} &           GT &  62.7 &   84.9 &   95.5 &         100.0 & 85.3 & 96.1 & 97.1 \\
          $\mH$~\cite{Homayounfar2017SportsFieldLocalization} &       \xmark &  54.1 &   82.9 &   92.4 &         100.0 & 81.8 & 100. & 100. \\
                                       \citet{chen2019sports} &           GT &  61.2 &   82.5 &   90.6 &         100.0 & 81.8 & 95.2 & 97.3 \\
                                                    DLT Lines &           GT &  39.2 &   57.4 &   72.1 &          89.8 & 60.3 & 82.6 & 96.5 \\

                                                      
\midrule

                    \rowcolor{OliveGreen!10} \acrshort{method} &                                    Pred &  38.8 &   69.1 &   89.4 &         100.0 & 73.3 & 95.3 & 96.6\\
                                        \citet{chen2019sports} &                   \cite{chen2019sports} &  35.8 &   66.3 &   84.4 &         100.0 & 69.5 & 94.6 & 96.3\\
                      \citet{Jiang2020OptimizingLearnedErrors} & \cite{Jiang2020OptimizingLearnedErrors} &  36.9 &   62.9 &   81.5 &         100.0 & 67.1 & 95.2 & 97.1\\
                                        \citet{chen2019sports} &                                    Pred &  28.8 &   58.0 &   77.3 &         100.0 & 62.1 & 91.7 & 94.9\\
                                                     DLT Lines &                                    Pred &  31.4 &   55.9 &   71.9 &          87.6 & 58.4 & 83.7 & 95.4\\

\citet{cioppa2021camera}                      & \cite{cioppa2021camera}           & \textcolor{gray}{\xmark} & \textcolor{gray}{\xmark} & \textcolor{gray}{\xmark} & 100. & \textcolor{gray}{\xmark} & \textcolor{gray}{88.5} & \textcolor{gray}{92.3} \\
\citet{Sha2020EndEndCamera}                   & \cite{Sha2020EndEndCamera}        & \textcolor{gray}{\xmark} & \textcolor{gray}{\xmark} & \textcolor{gray}{\xmark} & 100. & \textcolor{gray}{\xmark} & \textcolor{gray}{93.2} & \textcolor{gray}{96.1} \\
\citet{chu2022sports}                         &  \cite{chu2022sports}             & \textcolor{gray}{\xmark} & \textcolor{gray}{\xmark} & \textcolor{gray}{\xmark} & 100. & \textcolor{gray}{\xmark} & \textcolor{gray}{96.0} & \textcolor{gray}{97.0} \\
\citet{Shi2022SelfSupervisedShape}            & \cite{Shi2022SelfSupervisedShape} & \textcolor{gray}{\xmark} & \textcolor{gray}{\xmark} & \textcolor{gray}{\xmark} & 100. & \textcolor{gray}{\xmark} & \textcolor{gray}{96.6} & \textcolor{gray}{97.8} \\ 
\bottomrule
\end{tabularx}

%% file: tables/radial_lens_distortion.tex
\begin{tabularx}{\linewidth}{X|lcc|rrrr}
\toprule
\multirow{2}{*}{\textbf{Dataset}} & \multirow{2}{*}{\textbf{Seg.}} & \multirow{2}{*}{\textbf{$\tau$}} & \multirow{2}{*}{\textbf{LD}} & \multicolumn{3}{c}{\textbf{Accuracy@} [\%]} & \multirow{2}{*}{\textbf{CR}} \\
 &                               &                               &                               & \multicolumn{1}{c}{\textbf{5}} & \textbf{10} &  \textbf{20}                      &   \\ \midrule  \midrule

\multirow{4}{*}{SN-Calib-valid-center} & \multirow{2}{*}{GT} & 0.019 &      \xmark &  66.0 &   86.1 &   95.5 &       92.3 \\
          &    & 0.019 &  \checkmark &  \textcolor{OliveGreen}{66.6} &   85.0 &   94.7 &       \textcolor{BrickRed}{78.3} \\
\cline{2-8}
          & \multirow{2}{*}{Pred} & 0.019 &      \xmark &  54.9 &   79.9 &   92.3 &       92.9 \\
          &    & 0.019 &  \checkmark &  \textcolor{OliveGreen}{56.2} &   79.9 &   92.1 &       \textcolor{BrickRed}{86.2} \\
\cline{1-8}
\cline{2-8}
\multirow{2}{*}{WC14-test} & \multirow{2}{*}{GT} & $\infty$ &      \xmark &  64.4 &   86.7 &   96.0 &      100.0 \\
          &    & $\infty$ &  \checkmark &  \textcolor{OliveGreen}{68.4} &   87.3 &   95.5 &      100.0 \\

\bottomrule
\end{tabularx}